\documentclass[conference]{IEEEtran}
\IEEEoverridecommandlockouts

\usepackage{amsmath}
\usepackage{amssymb}
\usepackage{algorithm}
\usepackage{algorithmic}
\usepackage{graphicx}
\usepackage{bibarts}
\usepackage{cite}
\usepackage{subfigure}
\usepackage{diagbox}
\usepackage{caption}
\usepackage{color}

\begin{document}

\title{A Joint Gradient and Loss Based Clustered Federated Learning Design}

\author{
    \IEEEauthorblockN{
    Licheng Lin\IEEEauthorrefmark{1}, 
    Mingzhe Chen\IEEEauthorrefmark{1}\IEEEauthorrefmark{2}, 
    Zhaohui Yang\IEEEauthorrefmark{3},
    Yusen Wu\IEEEauthorrefmark{2},
    Yuchen Liu\IEEEauthorrefmark{4},
    }
    \IEEEauthorblockA{
    $\IEEEauthorrefmark{1}$Department of Electrical and Computer Engineering,
    University of Miami, Coral Gables, FL, 33146, USA
    \\
    $\IEEEauthorrefmark{2}$Frost Institute for Data Science and Computing, University of Miami, Coral Gables, FL, 33146, USA,
    \\
    $\IEEEauthorrefmark{3}$College of Information Science and Electronic Engineering, Zhejiang University, Hangzhou,  310027, China, 
    \\
    $\IEEEauthorrefmark{4}$Department of Computer Science, North Carolina State University, Raleigh, NC, 27695, USA
    \\
    Emails: \{lxl1293, mingzhe.chen, yxw1259\}@miami.edu,  
    yang\_zhaohui@zju.edu.cn,
    yuchen.liu@ncsu.edu 
    }
}

\maketitle

\begin{abstract}
    In this paper, a novel clustered FL framework that enables distributed edge devices with non-IID data to independently form several clusters in a distributed manner and implement FL training within each cluster is proposed. 
    In particular, our designed clustered FL algorithm must overcome two challenges associated with FL training. 
    First, the server has limited FL training information (i.e., the parameter server can only obtain the FL model information of each device) and limited computational power for finding the differences among a large amount of devices. 
    Second, each device does not have the data information of other devices for device clustering and can only use global FL model parameters received from the server and its data information to determine its cluster identity, which will increase the difficulty of device clustering. 
    To overcome these two challenges, we propose a joint gradient and loss based distributed clustering method in which each device determines its cluster identity considering the gradient similarity and training loss. The proposed clustering method not only considers how a local FL model of one device contributes to each cluster but also the direction of gradient descent thus improving clustering speed. By delegating clustering decisions to edge devices, each device can fully leverage its private data information to determine its own cluster identity, thereby reducing clustering overhead and improving overall clustering performance. 
    Simulation results demonstrate that our proposed clustered FL algorithm can reduce clustering iterations by up to $99$\% compared to the existing baseline.
\end{abstract}


\section{Introduction}\label{Introduction}
The development of mobile devices and video streaming applications (i.e., metaverse and virtual reality) motivates the development of distributed learning frameworks where devices can train their models locally using their own data. Federated learning (FL) \cite{mcmahan2017communication} is a such decentralized learning algorithm that allows devices to collaboratively learn a shared machine learning (ML) model while keeping their data localized on their own devices.
However, standard FL may not be applied for devices with non independent and identically distributed (non-IID) data since a standard FL method directly aggregates the ML models of devices without considering the data distributions of devices. To address this problem, one promising solution is to cluster the devices according to their data distributions such that the devices in a cluster with similar data distributions can collaboratively train a ML model thus solving the non-IID problem and improving training performance. However, designing clustered FL algorithms still presents several challenges including: 
1) The parameter server (PS) has limited information (i.e., FL model parameters) to determine cluster identities of all devices.
2) The PS has limited computational resource to identify differences among a large number of devices. 

Recently, a number of existing works such as in \cite{ghosh2019robust, briggs2020federated, sattler2020clustered, ghosh2020efficient, sattler2020byzantine, khan2020self, albaseer2021client, kim2021dynamic, feng2022mobility} have studied the design and deployment of clustered FL over wireless networks. 
In particular, the authors in \cite{ghosh2019robust} designed a clustered FL algorithm that first trains local models on each device, and then uses clustering algorithms such as k-means to cluster devices according to their locally trained convergent models. 
The work in \cite{briggs2020federated} developed a FL algorithm with hierarchical clustering approach. The designed algorithm first trains a global model over several FL training iterations and then clusters devices according to the similarities between updated local FL models.
The authors in \cite{sattler2020clustered} designed a clustered FL framework in which an original cluster containing all devices is recursively divided into smaller sub-clusters. The device clustering starts when the FL models are stationary and ends when the gradient norm of any devices in the sub-cluster is below a preset threshold value.
The work in \cite{ghosh2020efficient} designed a novel clustered FL which integrates the clustering algorithm into the training procedure, and to iteratively adjust the devices' cluster identities through FL process.
In \cite{sattler2020byzantine}, the authors investigated clustered FL under Byzantine attacks and shows that clustered FL can reliably detect and remove malicious clients. 
The authors in \cite{khan2020self} introduced a clustering algorithm based on social awareness for clustered FL and developed a heuristic algorithm to minimize the training time per FL iteration. Meanwhile, the designed clustering method in \cite{khan2020self} can eliminate the need of a centralized PS.
The work in \cite{albaseer2021client} designed a device selection approach for clustered FL to accelerate the convergence rate.
In \cite{kim2021dynamic}, a three-phased clustering algorithm based on generative adversarial network is introduced. The designed clustering method can create dynamic clusters and change the number of clusters over different iterations.
However, most of these existing works \cite{ghosh2019robust, briggs2020federated, sattler2020clustered, ghosh2020efficient, sattler2020byzantine, khan2020self, albaseer2021client, kim2021dynamic, feng2022mobility} focused on the design of centralized clustering methods which may lead to significant communication and computational overhead. Meanwhile, these works \cite{ghosh2019robust, briggs2020federated, sattler2020clustered, ghosh2020efficient, sattler2020byzantine, khan2020self, albaseer2021client, kim2021dynamic, feng2022mobility} considered the use of only local loss values of edge devices for device clustering without using other information (i.e., gradient vectors) of FL training.    

The main contribution of this paper is a novel clustered FL framework that enables distributed edge devices with non-IID data to independently form several clusters in a distributed manner and implement FL training within each cluster. In particular, our designed clustered FL algorithm must overcome two challenges associated with FL training. First, the server has limited FL training information (i.e., the PS can only obtain the FL model information of each device) and limited computational power for finding the differences among a large amount of devices. 
Second, each device does not have the data information of other devices for device clustering and can only use global FL model parameters received from the server and its data information to determine its cluster identity, which will increase the difficulty of device clustering. 
To overcome these two challenges, we propose a joint gradient and loss based distributed clustering method in which each device determines its cluster identity considering the gradient similarity and training loss. The proposed clustering method not only considers how a local FL model of one device contributes to each cluster but also the direction of gradient descent thus improving clustering speed. By delegating clustering decisions to edge devices, each device can fully leverage its private data information to determine its own cluster identity, thereby reducing clustering overhead and improving overall clustering performance. 
Simulation results over multiple datasets demonstrate that our proposed clustered FL algorithm can reduce the iterations required to cluster the devices correctly by up to $99$\% compared to the existing baseline.

The rest of the paper is organized as follows. The proposed clustered FL algorithm is described in Section \ref{Proposed_Clustered_FL_System}. Simulation settings and results are introduced in Section \ref{Simulation Results and Analysis}. Conclusions are drawn in Section \ref{Conclusion}.

\begin{figure}[t]\centering 
    \includegraphics[width=0.48\textwidth]{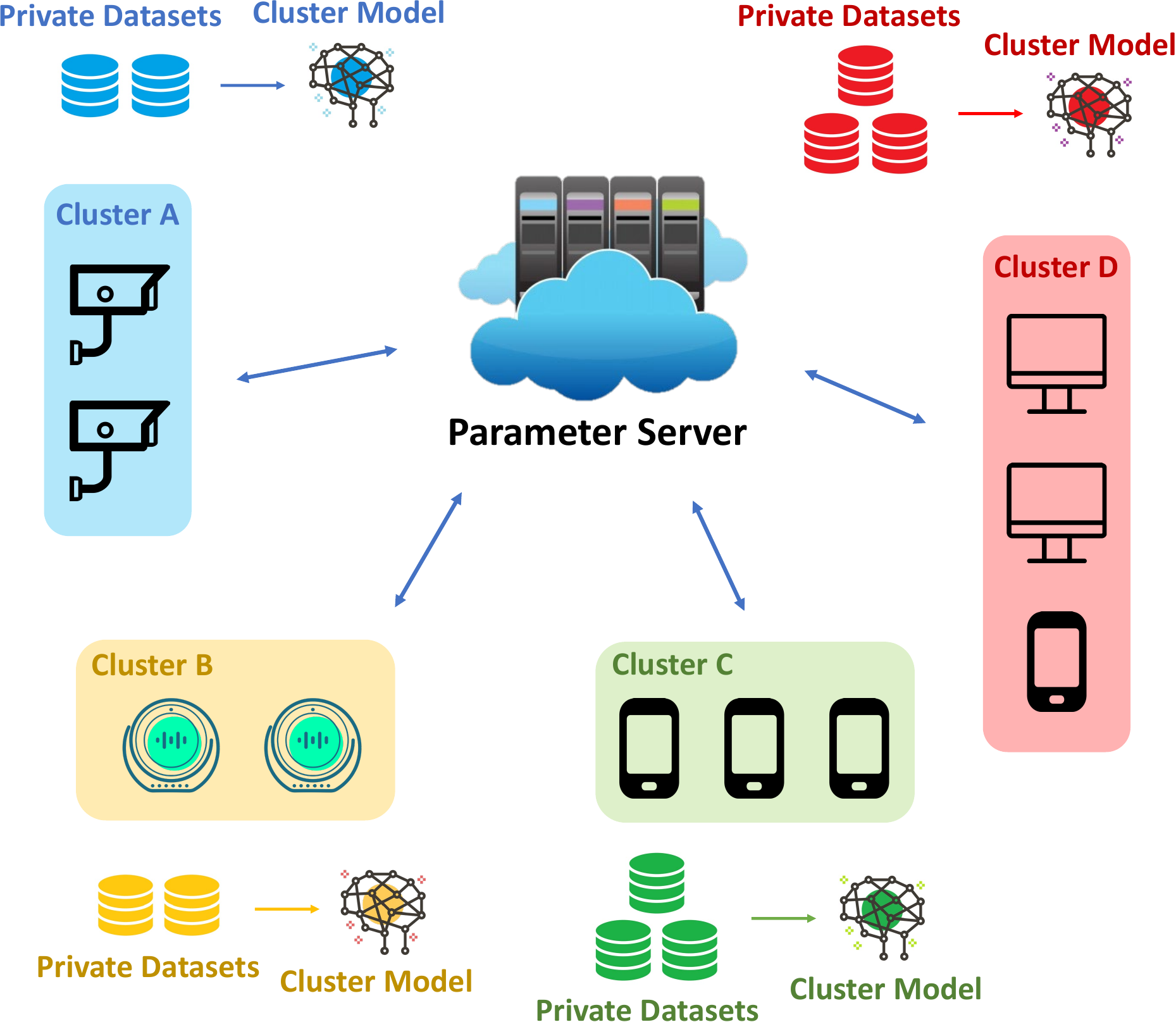}    
    \caption{A Framework of Clustered FL}    
    \label{fig:system}  
\end{figure}

\section{Proposed clustered FL System}\label{Proposed_Clustered_FL_System}

Consider a clustered federated learning framework in which one parameter server and a set $\mathcal{M}$ of $M$ devices collaboratively perform federated learning algorithms. 
In our model, devices have different datasets and hence the data distribution of the devices is non-IID. 
We assume that the total number of data distributions of all devices is $K$. 
To address the data heterogeneity problem \cite{zhao2018federated}, devices should be divided into $K$ clusters based on the characteristics of their datasets. 
The devices with similar data distributions are clustered into a group and jointly perform an FL training. 
In our model, we consider a general scenario where each device does not know the data distribution of other devices and the PS also does not know the data distributions of all devices. 
Hence, the PS cannot directly determine the cluster of each device and each device must use its limited FL parameter information to determine its cluster. 
To this end, it is necessary to design a novel clustered FL method where each device exploits its FL parameter information to determine its cluster individually.  
Next, we introduce our designed clustered FL algorithm. 
In particular, we first discuss the general process of clustered federated learning and then provide more details about the proposed clustering algorithm in clustered FL.

\subsection{General Procedure of Clustered FL}

Here, we introduce the general training process of clustered FL, which is summarized as follows:

\begin{enumerate}
    \item The server broadcast the parameters of $K$ FL models to all devices. We assume that $w^{t}_k$ represents the FL model parameters of cluster $k$ at iteration $t$. Here, the set of devices at each group $k$ may be changed according to the clustering results.
    \item Each device $i\in\mathcal{M}$ determines its cluster identity, i.e., which cluster it belongs to, via its private dataset and the model parameters received from the PS. Since this cluster identity would change through FL process, we denote the cluster identity of device $i$ at $t$-th iteration as $s_i^{t}$. Given its cluster identity $s_i^{t}$, each device will update its local FL model and transmit its FL parameters and cluster identity to the PS.
    \item The PS will aggregate the FL parameters with the same cluster identity and generate a global FL model. Since the devices are divided into $K$ clusters, the PS will generate $K$ global FL models.
    \item Repeat Steps 1-3 until converge. 
\end{enumerate}

From the training process of clustered FL, we see that clustered FL requires each device to use only its dataset and global FL models received from the PS to identify cluster identities and each device does not know the data distribution and cluster identify. Devices need to determine their clustering identities per iteration. 

\subsection{Proposed Clustered FL Algorithm}

Given the general process of clustered FL, in this subsection, we introduce our proposed clustered FL, which also consists of four steps: \textit{1)} cluster FL model broadcast, \textit{2)} device cluster identity determination, \textit{3)} local FL model update, and \textit{4)} local FL model aggregation, which are specified as follows.

\subsubsection{Cluster model broadcast}
Since the devices are grouped into $K$ clusters, the server will generate $K$ initial global FL models for all clusters. 
Hence, to implement our proposed clustered FL, the server will first broadcast the parameters of $K$ global FL models $\{\boldsymbol{w}_{1}^{t}, \boldsymbol{w}_2^{t}, \hdots, \boldsymbol{w}_K^{t}\}$ to the devices.

\subsubsection{Determination of cluster identity for each device}
Given the training process of clustered FL, two challenge must be solved when we design the device clustering algorithm. 
First, the device clustering method must be distributed since the server has limited FL training information (i.e., the PS can only obtain the FL model information of each device) and limited computational power for finding the differences among a large amount of devices. 
Second, each device does not have the data information of other devices for device clustering and can only use global FL model parameters received from the server and its data information to determine its cluster identity, which will increase the difficulty of device clustering. 
To overcome these two challenges, we propose a joint gradient and loss based distributed clustering method that consists of four steps: 1) Loss calculation, 2) Back-propagation, 3) Similarity calculation, and 4) Cluster identity determination, which are specified as follows:

\textbf{Step 1: Loss calculation} Given the parameters of $K$ FL models, $\{\boldsymbol{w}_1^{t}, \boldsymbol{w}_2^{t}, \hdots, \boldsymbol{w}_K^{t}\}$, device $i$ first calculates the loss with respect to each global FL model using a mini-batch of local data samples $\mathcal{Z}_i^{t}$, as follows:
\begin{equation}\label{eq:training_loss}
    \mathcal{L}_{i, k}^{t}(\mathcal{Z}_i^{t})=\sum_{\boldsymbol{z}\in \mathcal{Z}_i^{t}} l(\boldsymbol{w}_k^{t}, \boldsymbol{z}), \forall k=1, 2, \ldots, K.
\end{equation}
where $\boldsymbol{z}$ is a single sample in $\mathcal{Z}_i^{t}$, and $l(\boldsymbol{w}_k^{t}, \boldsymbol{z})$ is the loss value of model $\boldsymbol{w}_k^{t}$ with data sample $\boldsymbol{z}$.

\textbf{Step 2 Back-propagation: } Next, device $i$ can calculate the gradients of $K$ FL models based on the loss values obtained in the first step via back-propagation algorithm. In particular, we assume that the gradient of loss function $\mathcal{L}_{i, k}^{t}(\mathcal{Z}_i^{t})$ with respect to the global FL model $\boldsymbol{w}_k^{t}$ at device $i$ is $\nabla\mathcal{L}_{i, k}^{t}(\mathcal{Z}_i^{t}), \forall k=1, 2, \ldots, K$

\textbf{Step 3 Similarity calculation: } The gap between the global FL model $\boldsymbol{w}_k^{t}$ of cluster $k$ at iteration $t$ and the global FL model $\boldsymbol{w}_k^{t-1}$ of cluster $k$ at iteration $t-1$ is 
\begin{equation}\label{eq:Deltaw}
    \Delta\boldsymbol{w}_k^{t-1} = \boldsymbol{w}_k^{t} - \boldsymbol{w}_k^{t-1},
\end{equation}
In (\ref{eq:Deltaw}), $\Delta\boldsymbol{w}_k^{t-1}$ is the average gradient of all devices in cluster $k$ at iteration $t-1$.
The similarity between the local gradient $\nabla\mathcal{L}_{i, k}^{t}(\mathcal{Z}_i^{t})$ and $\Delta\boldsymbol{w}_k^{t-1}$ is calculated by
\begin{equation}\label{eq:gradient_similarity}
    S_{i, k}^{t}=\frac{\left(\nabla\mathcal{L}_{i, k}^{t}(\mathcal{Z}_i^{t})\right)\cdot \Delta\boldsymbol{w}_k^{t-1}}{\vert\nabla\mathcal{L}_{i, k}^{t}\vert\vert\Delta\boldsymbol{w}_k^{t-1}\vert}, \forall k=1, 2, \ldots, K.
\end{equation}
In (\ref{eq:gradient_similarity}), we use cosine similarity to characterize the similarity between local gradient and the latest global FL model update, which ignores the magnitude of gradient values and focuses on the direction of gradient descent. 
We can also use other functions to characterize the similarity between local gradient and the latest global FL model update. For example, if we consider both the magnitude and direction, we can use euclidean metric and treat the inverse of distance as similarity (i.e. $S_{i, k}^{t}=-\boldsymbol{d}(\nabla\mathcal{L}_{i, k}^{t},\Delta\boldsymbol{w}_k^{t-1}) =-\vert\vert\nabla\mathcal{L}_{i, k}^{t}-\Delta\boldsymbol{w}_k^{t-1}\vert\vert, \forall k=1, 2, \ldots, K.$)

\textbf{Step 4 Cluster identity determination: } Given (\ref{eq:gradient_similarity}), the cluster identity is estimated by
\begin{equation}\label{eq:combine_similarity_loss}
    s_i^{t}=\underset{k=1, 2, \ldots, K}{\mathrm{argmax}} \left(\lambda S_{i, k}^{t} + (1-\lambda) (-\mathcal{L}_{i, k}^{t})\right).
\end{equation}
where $\lambda$ is a weight parameter that controls the importance of the gradient similarity and the training loss for the cluster identification.
From (3), we see that the cluster identify of each device depends on the gradient similarity and the training loss.

\subsubsection{Local model update}
Given cluster identity $s_i^{t}$, device $i$ updates its local model as 
\begin{equation}\label{eq:local_update}
    \boldsymbol{w}_{i}^{(t+1)} = \boldsymbol{w}_{s_i^{t}}^{t} - \alpha\nabla\mathcal{L}_{i, s_i^{t}}^{t},
\end{equation}
where $\alpha$ is the learning rate. Then, device $i$ transmits its updated FL model parameters $\boldsymbol{w}_{i}^{(t+1)}$ and the cluster identity $s_i^{t}$ to the PS.

\subsubsection{Local FL model aggregation}
The uploaded models with the same cluster identity $s_i^{t}$ are aggregated by the PS so as to generate a global model of cluster $s_i^{t}$. 
Denote the set of devices identified as cluster $k$ as 
\begin{equation}\label{eq:cluster_device_set}
    \mathcal{M}_k=\{i|i\in\mathcal{M}, s_i^{t}=k\}.
\end{equation}
The global model aggregation of cluster $k$ can be represented as 
\begin{equation}\label{eq:aggregation}
    \boldsymbol{w}_k^{(t+1)} = \frac{1}{\vert\mathcal{M}_k\vert}\sum_{i\in\mathcal{M}_k}\boldsymbol{w}_i^{(t+1)}
\end{equation}

The full procedure of our proposed clustered FL algorithm is summarized in Algorithm \ref{algo:clustering_federated_learning}.

\addtolength{\topmargin}{0.05in}

\begin{algorithm}[t]
    \caption{Proposed Clustered Federated Learning}
    \begin{algorithmic}[1]\label{algo:clustering_federated_learning}
        \STATE \textbf{Input: } number of clusters $K$, number of clustering iterations $T$, number of devices $M$, set of devices $\mathcal{M}$, learning rate $\lambda$, $K$ initial cluster models $\{\boldsymbol{w}_1^{(0)}, \boldsymbol{w}_2^{(0)}, \hdots, \boldsymbol{w}_K^{(0)}\}$.

        \FOR{$t = 0, 1, \hdots, T-1$}
            \STATE \underline{server:} broadcast $\{\boldsymbol{w}_1^{t}, \boldsymbol{w}_2^{t}, \hdots, \boldsymbol{w}_K^{t}\}$ to all devices.

            \FOR{\underline{device} $i\in \mathcal{M}$ in parallel}
                
                \FOR{$k = 1, 2, \hdots, K$}
                    \STATE Calculate loss $\mathcal{L}_{i, k}^{t}$ by (\ref{eq:training_loss}).
                    \STATE Obtain gradient $\nabla\mathcal{L}_{i, k}^{t}$ via back-propagation.
                    \STATE Calculate gradient similarity $S_{i, k}^{t}$ by (\ref{eq:gradient_similarity}).
                \ENDFOR
                \STATE Estimate cluster identity by (\ref{eq:combine_similarity_loss})
                \STATE Update the local model by (\ref{eq:local_update})
                \STATE Upload $s_i^{t}$ and $ \boldsymbol{w}_i^{t}$ to the server.
            \ENDFOR

            \STATE \underline{server:} aggregate received models of each cluster based on (\ref{eq:aggregation})

        \ENDFOR

        \STATE \textbf{Output: } global cluster models $\{\boldsymbol{w}_1^{t}, \boldsymbol{w}_2^{t}, \hdots, \boldsymbol{w}_K^{t}\}$ and the cluster identities of devices $\{s_1^{t}, s_2^{t}, \hdots, s_M^{t}\}$.
    \end{algorithmic}
\end{algorithm}

\subsection{Empty Cluster Problem in Cluster FL Training}
To implement the proposed clustered FL, we may also need to solve a problem where one cluster may not have any devices during the training of clustered FL. 
This is because all the devices in this cluster may be misclassified into other clusters. Although this scenario may not happen frequently, it will significantly reduce the performance of clustered FL. In particular, 
at one clustered FL training iteration, if one cluster does not have any devices, the FL model of this cluster will not be updated. 
When the FL model is not updated, the gradient vector calculated by each device may not be correct such that the device will not select the cluster without updated FL model in the following iterations. 
In consequence, the number of clusters considered in our algorithm will be reduced. 
To address this issue, we can randomly select $K$ devices and allocate one device to one cluster.
Here, $K$ devices can have the same data distributions and we only need to make sure that each cluster will have one device per FL training iteration.

\section{Simulation Results and Analysis}\label{Simulation Results and Analysis}

We consider the implementation of the proposed clustered FL for learning tasks:{1)} 10 class hand written digits identification (i.e., MNIST\cite{lecun1998gradient}), {2)} 10 class fashion product images classification (i.e. FashionMNIST\cite{xiao2017/online}), {3)} 10 class objects classification (i.e., CIFAR10\cite{krizhevsky2009learning})
, and {4)} 62 class hand written letters and digits identification (i.e., EMNIST\cite{cohen2017emnist})
.
To evaluate the performance, we run our proposed clustered FL method in 4 experiments, each of which extracts its cluster task datasets from a classification dataset.
For comparison purpose, we use the iterative clustered FL scheme from \cite{ghosh2020efficient} as baseline.

\begin{figure*}[t]\centering 
    \includegraphics[width=\textwidth]{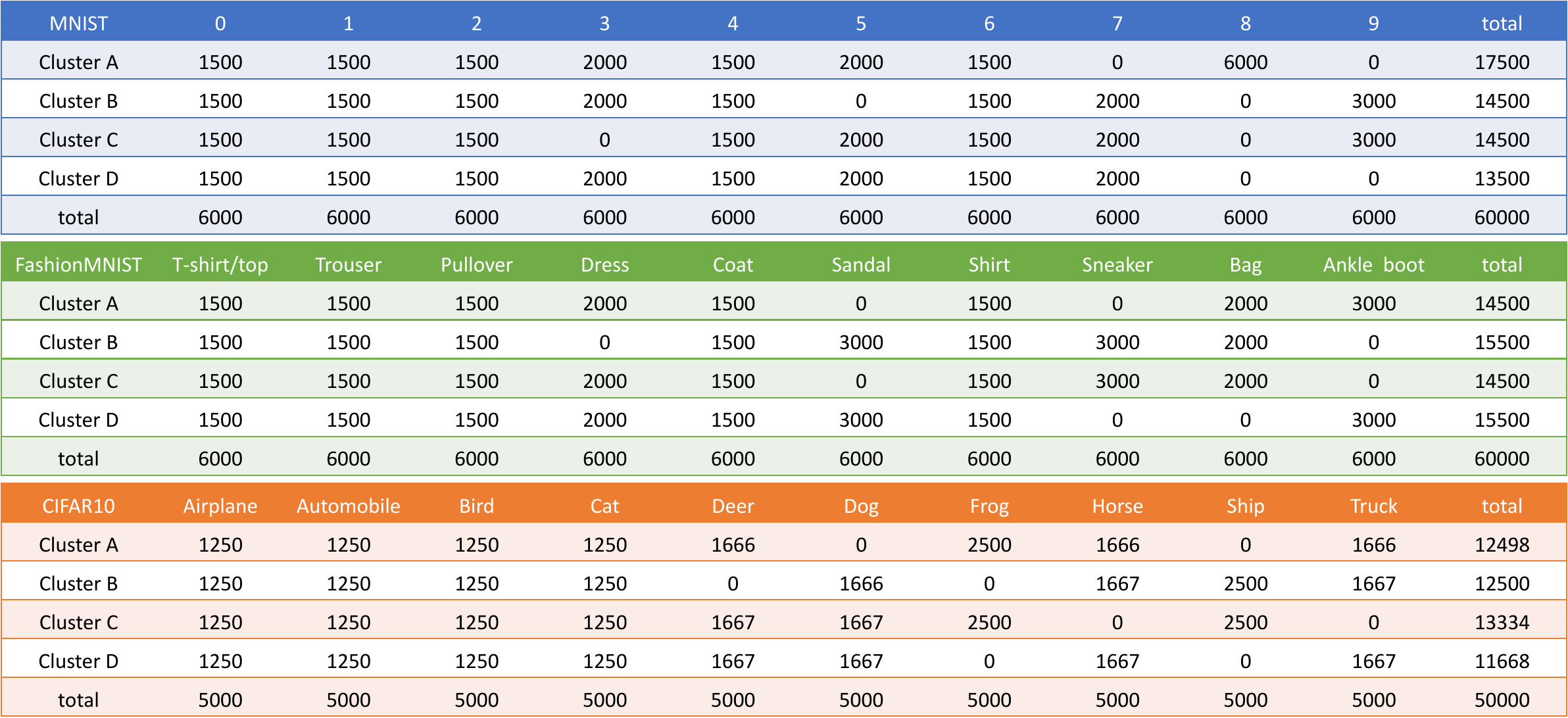}    
    \caption{Example splits of MNIST, FashionMNIST, and CIFAR10}    
    \label{tb:dataset_split}  
\end{figure*}

\subsection{Simulation Settings and Performance Metrics}

Here, we first explain how to generate the dataset for the devices in each cluster of each learning task. Then, we introduce the local FL model settings for each learning task. Finally, we describe the performance metrics used in the simulations. 

\subsubsection{Dataset Settings}
For the experiments on MNIST, FashionMNIST, and CIFAR10, we consider 80 devices jointly implement the clustered FL algorithm. These devices are equally divided into 4 clusters (i.e., clusters A, B, C, and D as shown in Fig. \ref{tb:dataset_split}) and each cluster has 20 devices. Each dataset totally has 10 class data and a device in each cluster has 8 class data. Hence the devices in different clusters will have at least 6 overlapped class data. In Fig. \ref{tb:dataset_split}, we show the data distribution of the four clusters for each learning task. From this figure, we see that, in MNIST, the devices in cluster A have a total of 17500 samples of number 0, 1, 2, 3, 4, 5, 6, 8, while the devices in cluster B have 14500 samples of number 0, 1, 2, 3, 4, 6, 7, 9. Hence, there are 6 overlapped class data between devices in cluster A and B.
The data samples of each cluster will be further distributed to its devices equally and randomly.

For EMNIST learning task, we consider the clustered FL is implemented by 200 devices which are divided into 8 clusters. 
To reduce ambiguity between the uppercase and lowercase forms of some easily hard-to-distinguish letters, we merged the uppercase and lowercase classes for the letters C, I, J, K, L, M, O, P, S, U, V, W, X, Y and Z, such that 62 class data are changed into 47 classes.
We further split these classes into clusters in the same manner as was done for the MNIST dataset, each of which has 40 class data.

\subsubsection{Learning Models}
\begin{figure}[t]\centering 
    \includegraphics[width=0.48\textwidth]{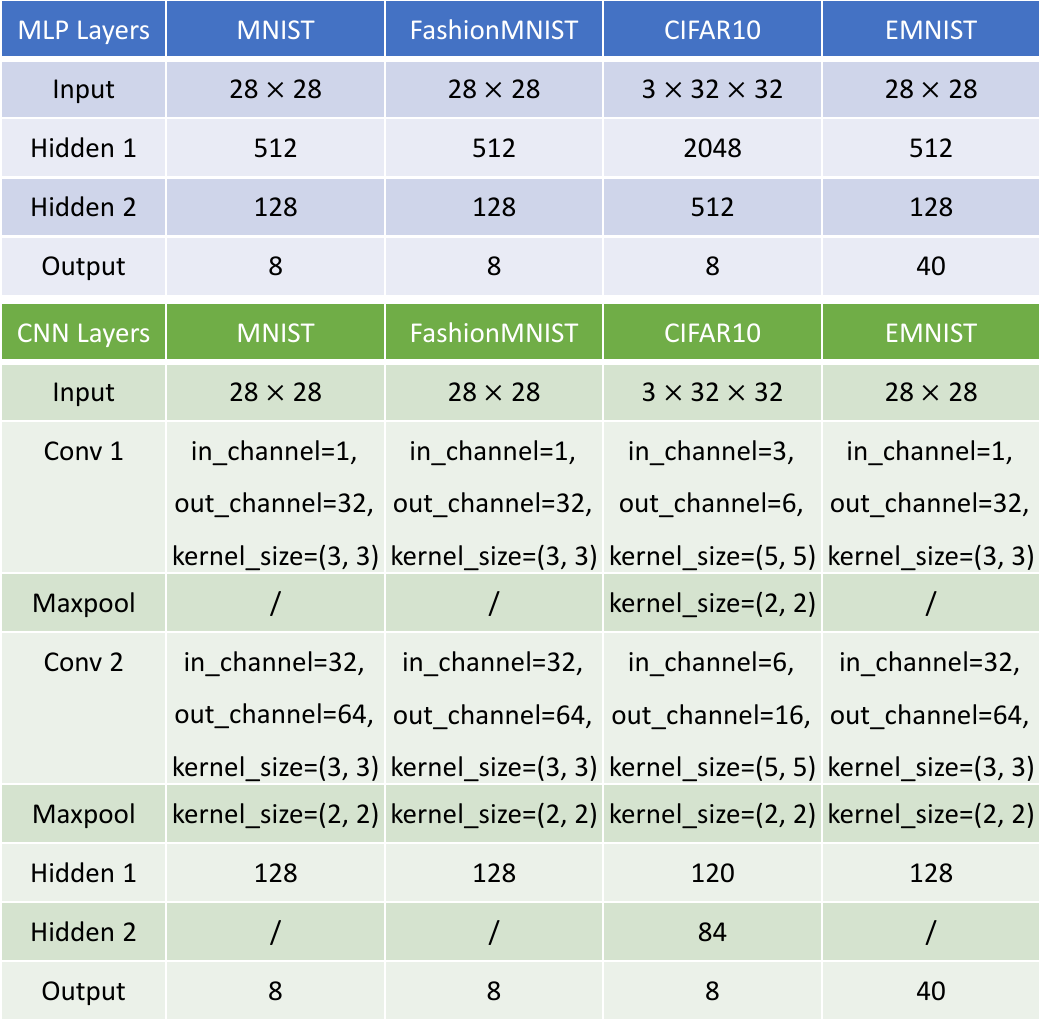}    
    \caption{Model architectures of the MLP and CNN for each experiment}    
    \label{tb:models}  
\end{figure}

For each learning task, we consider the use of two neural network models as local FL models. The first one is a multi-layer perceptron (MLP) with three fully-connected layers with ReLU activation. The second model is a convolutional neural network (CNN) which consists of two convolutional layers followed by fully-connected layers. The detailed MLP and CNN model architectures are shown in Fig.~\ref{tb:models}.

\subsubsection{Performance Metrics}

To measure the clustering accuracy of the clustered FL algorithm, we use purity which is defined as the percentage of devices that are classified correctly.
The purity $P^{t}$ at iteration $t$ is mathematically expressed as
$$P^{t}=\frac{1}{\vert\mathcal{M}\vert}\sum_{k}\max_{j}\vert\mathcal{M}_j^{*}\cap \mathcal{M}_k^{t}\vert,$$
where $\mathcal{M}_j^{*}$ is the ground truth set of devices at cluster $j$, and $\mathcal{M}_k^{t}$ is the set of devices that are clustered by the clustered FL algorithm at iteration $t$, with
$$\mathcal{M}_k^{t}=\{i|i\in\mathcal{M}, s_i^{t}=k\}, \forall k=1, 2, \hdots, K$$
is the set of devices with the same cluster identity $k$ at iteration $t$.

In order to demonstrate that proposed algorithm brings better performance to the clustered FL training, we also use test accuracy to measure the training effect of clustered FL. 
While splitting the training dataset, we also split the test dataset for each user which has the same sample distribution as their training dataset.
The total test accuracy of the clustered FL system is obtained by averaging test accuracy of all users.

\subsection{Results Analysis}

\begin{figure*}[t]
    \centering
    \subfigure[]
    {
        \includegraphics[width=0.33\textwidth]{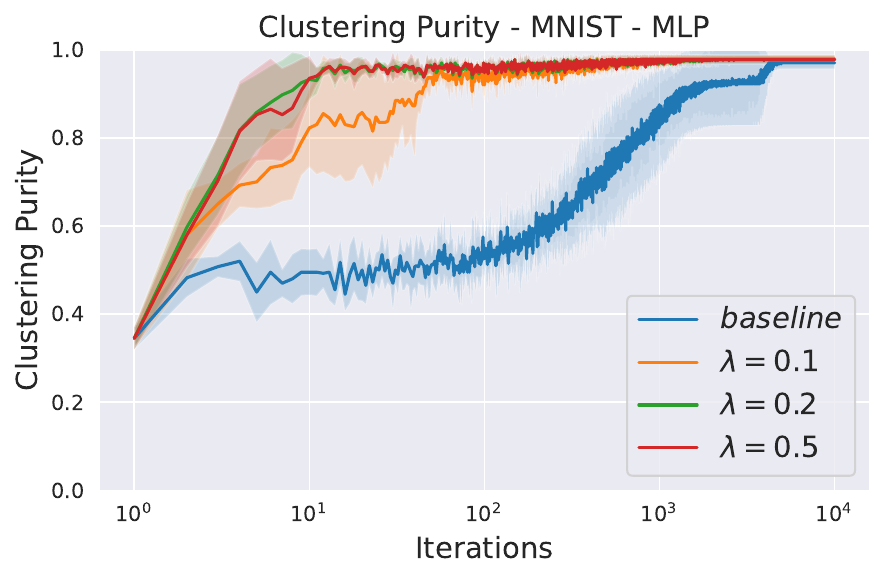}
        \label{fig:MNIST_Clustering_Purity_MLP}
    }
    \hspace{-6mm}
    \subfigure[]
    {
        \includegraphics[width=0.33\textwidth]{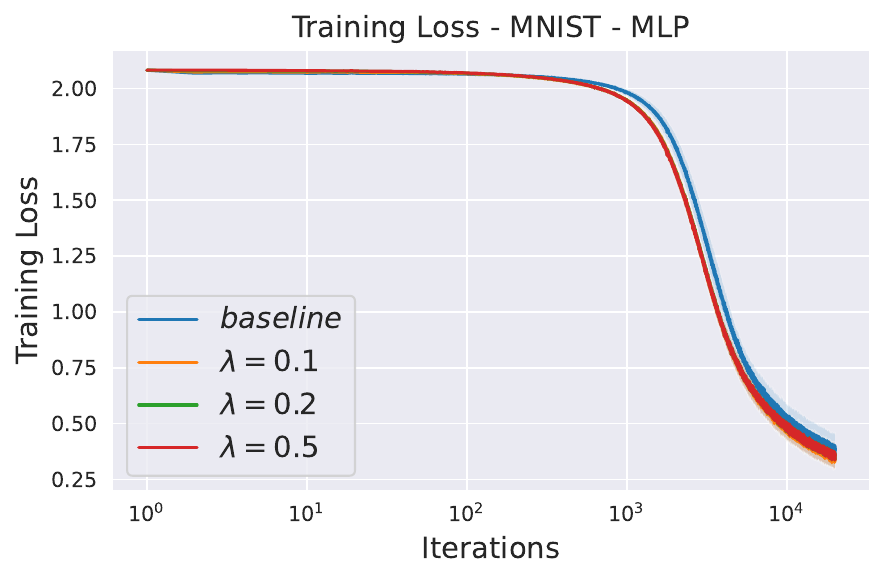}
        \label{fig:MNIST_Training_Loss_MLP}
    }
    \hspace{-6mm}
    \subfigure[]
    {
        \includegraphics[width=0.33\textwidth]{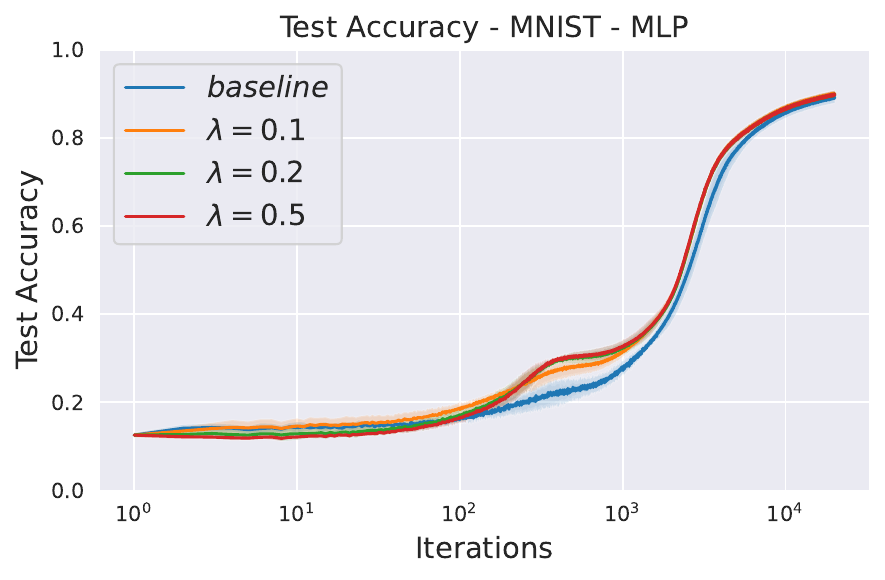}
        \label{fig:MNIST_Test_Accuracy_MLP}
    }
    \\
    \vspace{-2mm}
    \subfigure[]
    {
        \includegraphics[width=0.33\textwidth]{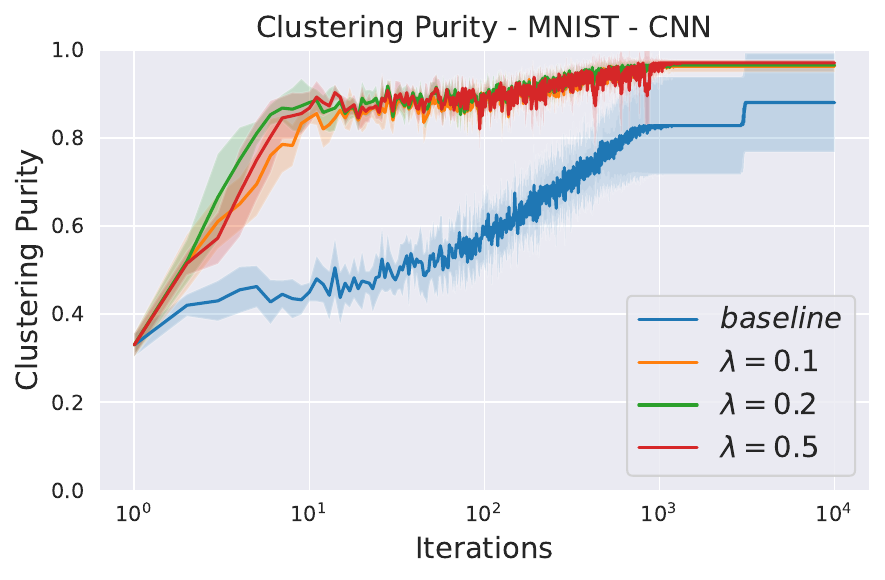}
        \label{fig:MNIST_Clustering_Purity_CNN}
    }
    \hspace{-6mm}
    \subfigure[]
    {
        \includegraphics[width=0.33\textwidth]{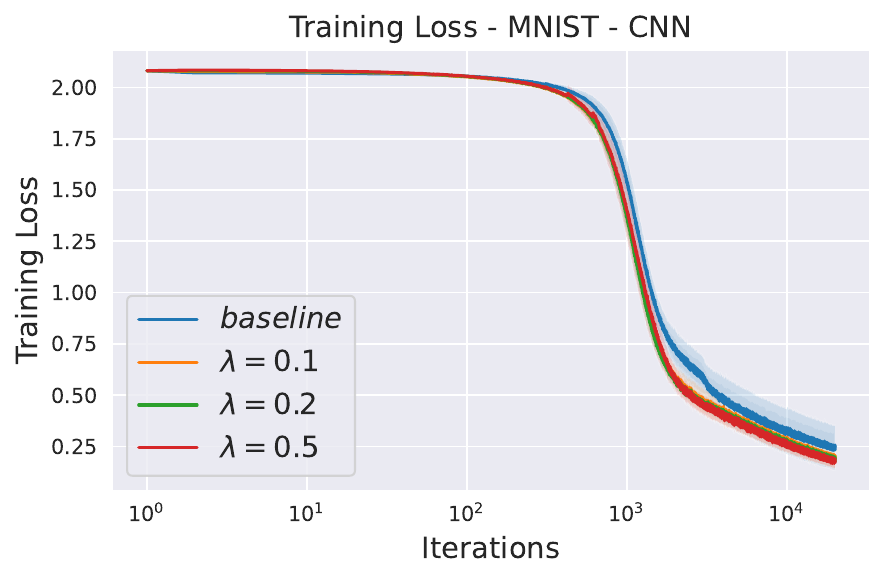}
        \label{fig:MNIST_Training_Loss_CNN}
    }
    \hspace{-6mm}
    \subfigure[]
    {
        \includegraphics[width=0.33\textwidth]{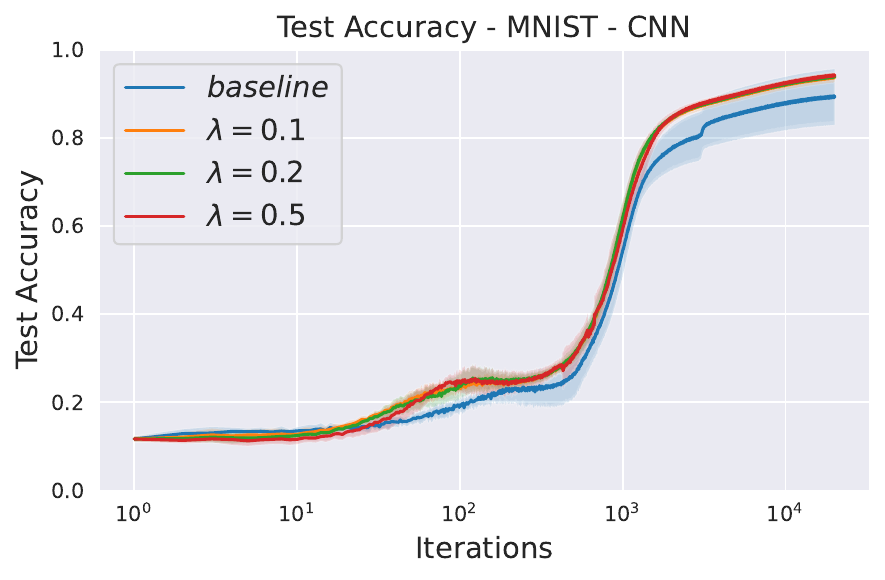}
        \label{fig:MNIST_Test_Accuracy_CNN}
    }
    \vspace{-2mm}
    \caption{Performance metrics vary as the number of clustered FL iterations changes on MNIST experiment.}
    \vspace{-2mm}
    \label{fig:exp_MNIST}
\end{figure*}

In Fig. \ref{fig:exp_MNIST}, we show how the clustering purity, training loss, and test accuracy vary as the number of training iterations changes. This experiment is implemented over MNIST dataset. 
Fig. \ref{fig:MNIST_Clustering_Purity_MLP}, Fig. \ref{fig:MNIST_Training_Loss_MLP}, and Fig. \ref{fig:MNIST_Test_Accuracy_MLP} are results of MNIST experiments where MLP is used as FL models, while Fig. \ref{fig:MNIST_Clustering_Purity_CNN}, Fig. \ref{fig:MNIST_Training_Loss_CNN}, and Fig. \ref{fig:MNIST_Test_Accuracy_CNN} are results of experiments where CNN is used as FL models. 
Figs. \ref{fig:MNIST_Clustering_Purity_MLP} and \ref{fig:MNIST_Clustering_Purity_CNN} show how clustering purity changes as the number of iterations increases.
From Fig. \ref{fig:MNIST_Clustering_Purity_MLP}, we see that the proposed algorithm with $\lambda=0.2$ can reduce $99$\% iterations to achieve $0.9$ clustering purity compared to the baseline. 
This is because the proposed clustered FL algorithm jointly uses gradient direction and loss value to cluster devices. 
From Fig. \ref{fig:MNIST_Clustering_Purity_MLP}, we also see that when $\lambda$ changes from $0.1$ to $0.5$, the proposed algorithm can achieve higher purity at the beginning.
This is because the gradient direction can cluster devices better than loss value at the beginning, therefore higher weight for gradient direction brings better performance.
Fig. \ref{fig:MNIST_Training_Loss_MLP} and Fig. \ref{fig:MNIST_Test_Accuracy_MLP} show that the proposed clustered FL algorithm can reduce $14$\% iterations to achieve $0.8$ test accuracy compared to the baseline.
Fig. \ref{fig:MNIST_Clustering_Purity_CNN}, Fig. \ref{fig:MNIST_Training_Loss_CNN}, and Fig \ref{fig:MNIST_Test_Accuracy_CNN} show that when CNN is used as FL model, the cluster performance and training efficiency of the proposed algorithm is also better, compared to the baseline. 
This stems from the fact that the clustering process of the proposed algorithm is more efficient, which accelerates the training of FL.

\begin{figure*}[t]
    \centering
    \subfigure[]
    {
        \includegraphics[width=0.33\textwidth]{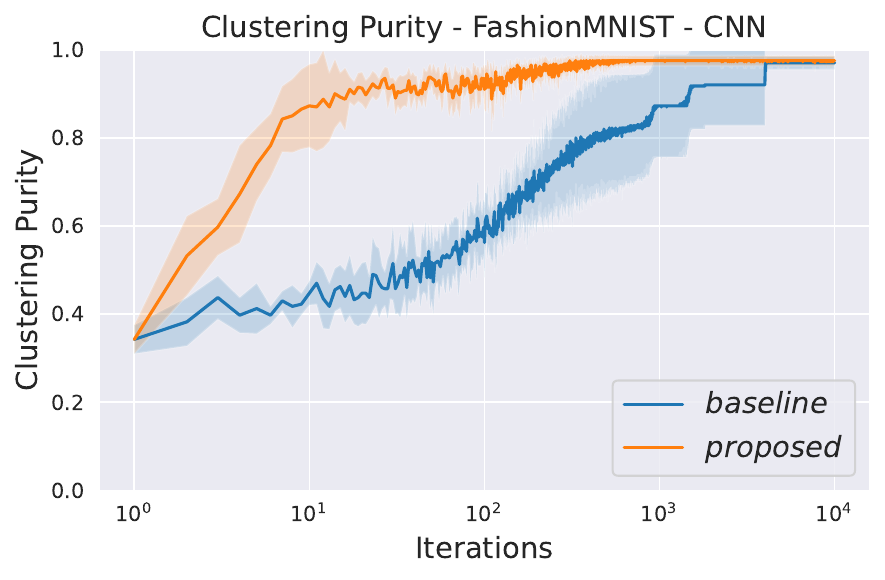}
        \label{fig:FashionMNIST_Clustering_Purity_CNN}
    }
    \hspace{-6mm}
    \subfigure[]
    {
        \includegraphics[width=0.33\textwidth]{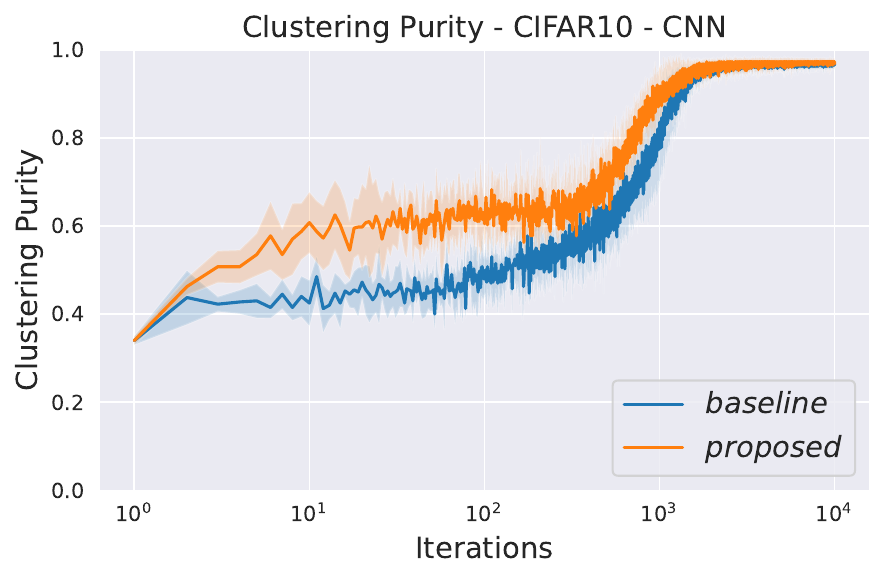}
        \label{fig:CIFAR10_Clustering_Purity_CNN}
    }
    \hspace{-6mm}
    \subfigure[]
    {
        \includegraphics[width=0.33\textwidth]{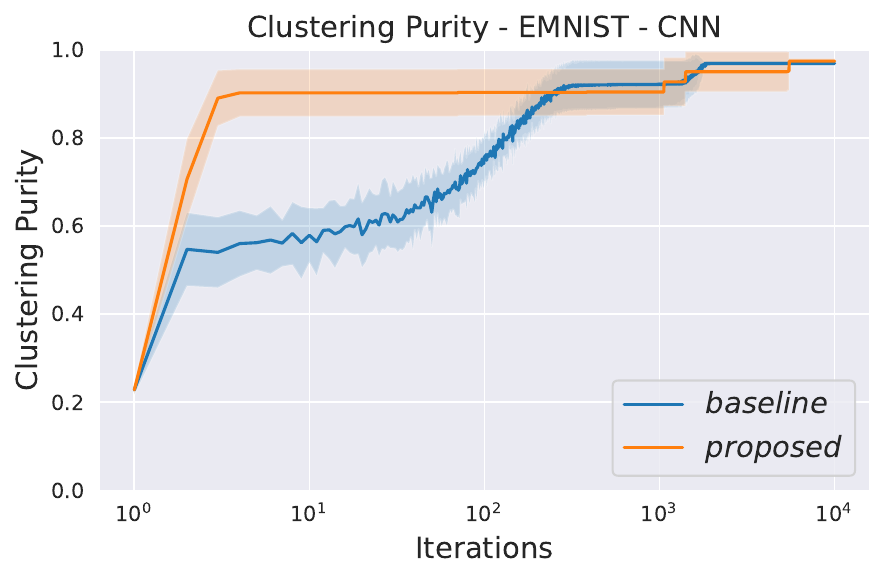}
        \label{fig:EMNIST_Clustering_Purity_CNN}
    }
    \vspace{-2mm}
    \caption{Clustering Purities vary as the number of iterations increases on FashionMNIST, CIFAR10, EMNIST experiment.}
    \vspace{-2mm}
    \label{fig:exp_FashionMNIST_CIFAR_EMNIST_CNN}
\end{figure*}

In Fig. \ref{fig:exp_FashionMNIST_CIFAR_EMNIST_CNN}, we show how the clustering purity varies as the number of training iteration increases in  experiments implemented over FashionMNIST, CIFAR10, and EMNIST.
From Fig. \ref{fig:FashionMNIST_Clustering_Purity_CNN}, Fig. \ref{fig:CIFAR10_Clustering_Purity_CNN}, and Fig. \ref{fig:EMNIST_Clustering_Purity_CNN}, we see that the proposed algorithm respectfully reduces iterations required to achieve $0.9$ clustering purity by up to $98$\%, $21$\% and $97$\% compared to the baseline.
This is because the proposed algorithm can jointly use gradient direction and loss value to cluster devices.

\section{Conclusion}\label{Conclusion}

In this paper, we have developed a novel clustered FL framework that enables distributed edge devices with non-IID data to independently form several clusters in a distributed manner and implement FL training within each cluster. 
In particular, our designed device method considered two unique FL features: 1) limited FL training information and computational power at the PS and 2) each device does not have the data information of other devices for device clustering and can only use global FL model parameters received from the server and its data information to determine its cluster identity. We have proposed a joint gradient and loss based distributed clustering method, in which each device determines its cluster identity considering the gradient similarity and training loss. 
The proposed clustering method not only considers how a local FL model of one device contributes to each cluster but also the direction of gradient descent thus improving clustering speed.  
Simulation results over multiple datasets demonstrate that our proposed clustered FL algorithm can yield significant gains compared to the existing method.
\def\baselinestretch{0.9}
\bibliography{ref}
\bibliographystyle{IEEEtran}

\end{document}